\documentclass[conference]{IEEEtran}
\IEEEoverridecommandlockouts
\usepackage{cite}
\usepackage{amsmath,amssymb,amsfonts}
\usepackage{algorithmic}
\usepackage{graphicx}
\usepackage{subfigure}
\usepackage{textcomp}
\usepackage{multirow}
\usepackage{booktabs}
\usepackage{xcolor}
\usepackage{hyperref}
\usepackage{amsmath, amsthm}
\usepackage[ruled,vlined,linesnumbered]{algorithm2e}
\theoremstyle{definition}
\newtheorem{definition}{Definition}
\newtheorem{theorem}{Theorem}
\newtheorem*{problem}{Problem Statement}
\def\BibTeX{{\rm B\kern-.05em{\sc i\kern-.025em b}\kern-.08em
    T\kern-.1667em\lower.7ex\hbox{E}\kern-.125emX}}
\begin{document}

\title{FoundAna: A GNN-assisted Foundation Model\\ for Graph Anomaly Detection
}

\author{\IEEEauthorblockN{
		Suprim Nakarmi\IEEEauthorrefmark{1},
		Chahana Dahal\IEEEauthorrefmark{1},
		Yue Zhao\IEEEauthorrefmark{2},
        Junggab Son\IEEEauthorrefmark{1} and
		Zuobin Xiong\IEEEauthorrefmark{1}}
		\IEEEauthorblockA{
		\IEEEauthorrefmark{1}Department of Computer Science, University of Nevada Las Vegas, Las Vegas, USA\\
		\IEEEauthorrefmark{2}Department of Computer Science, University of Southern California, Los Angeles, USA\\
		\IEEEauthorrefmark{1} {\it \{suprim.nakarmi, chahana.dahal, junggab.son, zuobin.xiong\}@unlv.edu};
		\IEEEauthorrefmark{2} {\it yue.z@usc.edu}}
		}

\maketitle

\begin{abstract}
Graph anomaly detection aims to identify graph structures (e.g., nodes, edges, or subgraphs) that deviate significantly from expected patterns, which supports critical applications in fraud detection, spam identification, network intrusion, etc. 
Despite the growing methods in the field, existing approaches follow a one-model-per-dataset paradigm, limiting their transferability across diverse real-world scenarios due to task heterogeneity, label scarcity, and domain variability. 
In this work, we introduce \textbf{FoundAna}, a GNN-assisted \underline{Found}ation Model for Graph \underline{An}om\underline{a}ly Detection -- the first foundation model framework designated for generalizable, cross-graph anomaly detection by combining GNNs and transformers. 
FoundAna integrates an anomaly detection-specific GNN component with a standard transformer encoder augmented by four complementary positional encodings, which enable the model to capture both local and global structural information. 
Specifically, the positional encoding enriched node representations are passed through attribute and adjacency decoders, and the reconstruction errors serve as the anomaly score. 
Extensive experiments on nine benchmark datasets spanning financial, social, and citation network domains demonstrate that FoundAna consistently outperforms state-of-the-art baselines. 
The code implementation and \underline{Supplementary materials} are here: \href{https://github.com/FoundAna331/FoundAna} {https://github.com/FoundAna331/FoundAna}.
\end{abstract}


\section{Introduction}
Anomaly Detection (AD) has been an important topic in the field of research and practice, which refers to identifying abnormal data samples (also called outliers) that deviate significantly from the distribution of the majority~\cite{grubbs1969procedures, chandola2009anomaly}. 
A wide spectrum of applications, including detecting fraudulent transactions in insurance, banking, etc.~\cite{chandola2009anomaly}; identifying malicious network intrusion in cybersecurity infrastructure~\cite{munir2018deepant}; and ensuring safety in aviation and industrial control systems~\cite{munir2019comparative,chandola2009anomaly}, heavily rely on such anomaly detection techniques. 
Classical approaches in AD employ statistical, distance-based, density-based, and kernel-based methods to characterize abnormality through mathematical criteria~\cite{munir2019comparative}.
However, these methods fail to scale to the complex, high-dimensional data distributions prevalent in modern applications, leading to the advent of deep learning-based models such as deep autoencoders, variational generative models, and self-supervised architectures~\cite{chalapathy2019deep}. 
Despite these advances, a critical limitation persists in almost all learning-based methods, as they are designed for Euclidean data modalities (e.g., images and tabular records), which are less effective in capturing topological dependencies inherent in graph-structured data~\cite{ma2021comprehensive}.

The semantic and topological complexity of graph-structured data is determined by the attributes of nodes and their structural connectivity within the graph. 
This dual nature induces a heterogeneous anomaly landscape that spans multiple levels: 
\textit{node anomaly} may represent a bot account generating thousands of spurious connections in a social network~\cite{wang2016botnet}; 
an \textit{edge anomaly} may correspond to a sudden, high-value financial transaction between accounts that rarely interact~\cite{zakaria2025detecting}; 
and a \textit{(sub)graph anomaly} may indicate a rare or structurally deviant molecular ring deviating from known chemical norms~\cite{wang2026anomaly}. 
In the traditional Graph Anomaly Detection (GAD), Graph Neural Networks (GNNs) are the dominant backbone, enabling joint encoding of structural and attribute information through message passing mechanisms~\cite{ma2021comprehensive}. 
However, GAD using GNNs faces three fundamental challenges: 
(i) task heterogeneity: node, edge, and subgraph-level anomaly detection require distinct model designs, 
(ii) label scarcity: anomalies are inherently rare and expensive to annotate in a graph at scale, 
and (iii) domain variability: graph characteristics differ across application domains, limiting cross-dataset transferability. 
These challenges have motivated a recent shift toward Graph Foundation Models (GFMs) for GAD, which uses a single and powerful framework to generalize across tasks and domains, with limited labeled data.

Foundation models~\cite{bommasani2021opportunities} are typically pre-trained on large and diverse data corpora and subsequently adapted to downstream tasks with minimal fine-tuning. 
Specifically, all of the modern FMs in the Natural Language Processing and Computer Vision domains use transformers as the backbone~\cite{zhou2025comprehensive, awais2025foundation}. 
Inspired by this success, the graph learning community has developed transformer-based GFMs that generalize across diverse graph tasks and domains~\cite{liu2025graph, wang2025graph}, which provide unbounded global self-attention to capture long-range dependencies, thereby overcoming the over-smoothing and over-squashing issues of message-passing GNNs~\cite{liu2023towards}.
\textit{However, leveraging standard transformers for graphs is non-trivial, as they treat nodes as an unordered set of tokens and lack any built-in mechanism to encode graph topology.} 
Therefore, the selection of positional encoding, which provides the structural information for node representations before attention, is of critical importance. 
For example, using Laplacian eigenvectors as positional encoding provides a mathematical map of the graph's global geometry, and a myriad of works have demonstrated graph positional encoding for general graph tasks in the literature~\cite{zhang2020graph, xia2024opengraph}.
Yet, very few studies have investigated positional encoding specific to anomaly detection methods. 

\begin{figure}[t]
	\subfigure[Node degree]{
			\centering
			\includegraphics[width=0.45\linewidth]{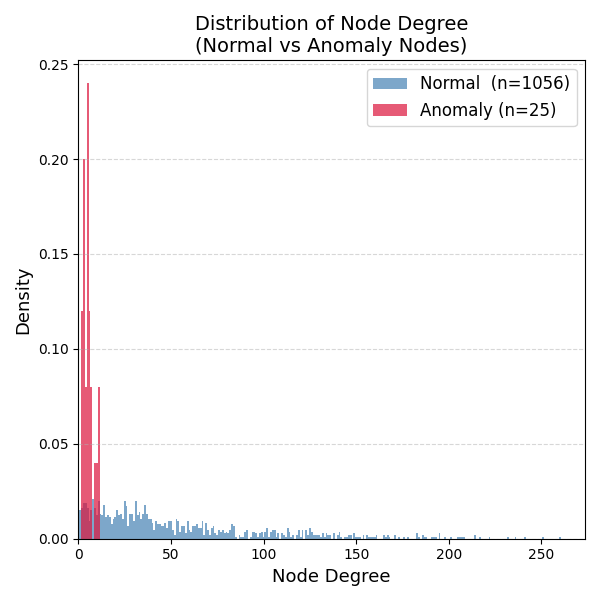}
			\label{fig:node_d}
	}
	\subfigure[Residual distance]{
			\centering
			\includegraphics[width=0.45\linewidth]{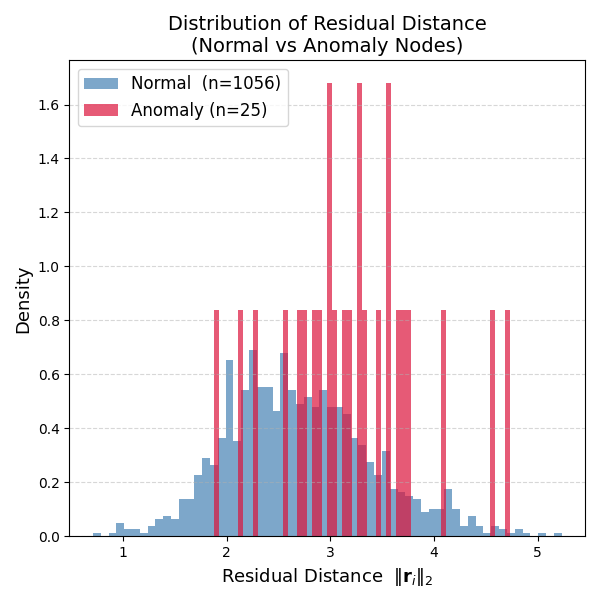}
			\label{fig:res_dist}
	}
	\caption{Distribution of node degree and residual distance in normal and anomaly nodes. The figure was plotted using the Facebook dataset.}  
	\label{fig:ano_vs_nor}
\end{figure}

In this work, we develop a GFM by integrating the benefit of GNNs for local structural learning (w.r.t. local anomaly) and the strength of the transformer for capturing the global graph attention (w.r.t. universal performance). 
Furthermore, we adopt four different positional encoding methods, including the Stable and Expressive Positional Encoding (SPE)~\cite{huang2024stability}, Random Walk Structural Encoding (RWSE)~\cite{rampavsek2022recipe}, node degree encoding, and residual distance encoding. 
In addition, the node degree and residual distance encoding are specialized to capture anomaly nodes with a distinct distribution from normal nodes, as shown in Figure~\ref{fig:node_d} and Figure~\ref{fig:res_dist}.
To the best of our knowledge, no prior work has proposed an anomaly-specific GNN coupled with transformers as the foundational graph anomaly detection model. 
The main contributions are summarized as follows:

\begin{itemize}

    \item We propose the first GNN-assisted transformer-based GFM specifically designed for GAD that generalizes well in cross-domain and cross-task detections.

    \item To echo the GNN-assisted transformer, we introduce a novel multi-scale positional encoding scheme integrating SPE, RWSE, residual distance, and node degree, providing both anomaly-relevant local structure and global context for transformer attention.

    \item Experimental results on nine datasets and multiple baselines, including GNNs and GFMs, confirm the superiority of the proposed method, especially in the few-shot learning scenario.
\end{itemize}

\section{Related Work}
\label{sec:related}

\subsection{Graph Anomaly Detection}
Early approaches to GAD relied on statistical or shallow learning methods; however, leveraging GNNs has gained much popularity for their structurally aware representations, which substantially improved detection performance~\cite{qiao2025deep, ma2021comprehensive}. 
Over time, GNN-based GAD has evolved into three primary directions: reconstruction-based, contrastive learning-based, and GNN aggregation-based. 
\textbf{Reconstruction-based} methods typically train models to encode normal graph patterns and detect anomalies as nodes or structures that are poorly reconstructed~\cite{roy2024gad}. 
\textbf{Contrastive-learning} based method provides a semi-supervised learning mechanism for GAD by pushing normal nodes to be similar to their local neighborhoods and pulling anomalies away~\cite{liu2021anomaly}. 
\textbf{GNN-aggregation} consists of a family of methods that addresses GAD through careful design of GNN aggregation and spectral filtering~\cite{tang2022rethinking}. 
All these methods are designed to detect anomalies for a specific data domain and a specific graph task, such as node, edge, and graph level.

Even though GAD models have superior performances, they follow \textit{a one model per dataset} setting, meaning a model trained on a fraud detection graph cannot be reused for a social network anomaly task without retraining, which leads to a common issue of heavy data annotation demands and poor generalization~\cite{liu2024arc, niu2024zero}.

\subsection{Graph Foundation Models}
\label{sub:fm}
Inspired by foundation models in NLP and computer vision, GFMs emerged to move from dataset-specific GNNs toward unified, transferable graph models. 
GraphBERT~\cite{zhang2020graph} introduced pure-attention graph learning, while self-supervised pretraining on molecular graphs demonstrated that graph transformers can learn transferable representations without task-specific supervision. 
One For All (OFA)~\cite{liu2024one} reformulated all graph tasks into a unified node-classification problem using prompt graphs. 
OpenGraph~\cite{xia2024opengraph} resolved feature heterogeneity through LLM-based shared embedding spaces, and GFT~\cite{wang2024gft} introduced a transferable tree-structured vocabulary for graph-agnostic learning. 

Building on these foundations, recent works have extended GFMs to graph anomaly detection in two directions: GNN-backbone and transformer-backbone. 
GNN-backbone approaches leverage message-passing architectures as the core encoder. 
UniGAD~\cite{lin2024unigad} unified node, edge, and graph-level AD under a single model, showing that multi-level joint training regularizes representations and improves performance at each graph-level task. 
UniFORM~\cite{song2025uniform} handles AD across diverse graph types under a single training objective, and GFM-UAD~\cite{han2026graph} extended this to a fully unified framework pretrained once and deployed across all three graph-level tasks. 
Similarly, AnomalyGFM~\cite{qiao2025anomalygfm} addresses cross-graph feature heterogeneity by encoding anomaly signals as residual distances, which is the deviation of a node's representation from its neighbors.
Transformer-backbone methods replace GNN encoders with attention mechanisms. 
This direction includes TFM4GAD~\cite{liu2026tabular}, which encodes structural position through Laplacian embeddings and adds PageRank-based structural characteristics and neighborhood aggregation using beta wavelet filters. 
UNIP~\cite{niu2024zero}, which designs transferable neighborhood prompts enabling zero-shot AD on entirely unseen graphs; 
and ARC~\cite{liu2024arc}, which reframes GAD as an in-context learning problem where a few labeled examples guide anomaly scoring without retraining.  


However, existing GFMs for anomaly detection function on fixed or pre-aligned feature spaces and do not fully leverage transformer architectures~\cite{lin2024unigad,liu2024arc,qiao2025anomalygfm}, which limits their ability to model both local neighborhood structure and global graph topology simultaneously. 
To address this gap, we propose FoundAna, which combines a GNN backbone with a transformer to jointly capture local and global structural context. 
The GNN encodes local neighborhood information, while the transformer leverages a combination of multi-scale positional encoding to model global structural position.

\section{Preliminaries}
In this section, we briefly conceptualize GFMs, GAD, and the problem formulation.




\subsection{Graph Anomaly Detection}
GAD aims to identify nodes, edges, or subgraphs that deviate significantly from the expected normal data samples within a graph. 
Here, we define the attributed graphs and GAD.

\begin{definition}[Attributed Graph]
An attributed graph is defined as $\mathcal{G} = (\mathcal{V}, \mathcal{E}, \mathbf{X})$, where $\mathcal{V} = \{v_1, v_2, \ldots, v_n\}$ is the set of nodes with $|\mathcal{V}| = n$, $\mathcal{E} \subseteq \mathcal{V} \times \mathcal{V}$ is the set of edges, and $\mathbf{X} \in \mathbb{R}^{n \times d}$ is the node feature matrix with input dimensionality $d$, where $\mathbf{x}_i \in \mathbb{R}^d$ denotes the feature vector of node $u_i$. The graph topology is encoded by an adjacency matrix $\mathbf{A} \in \{0,1\}^{n \times n}$, where:
$$
\mathbf{A}_{ij} = 
\begin{cases} 
1 & \text{if an edge exists between } v_i \text{ and } v_j, \\ 
0 & \text{otherwise.} 
\end{cases} 
$$
\end{definition}

\begin{figure*}[t]
    \centering
    \includegraphics[width=0.9\linewidth]{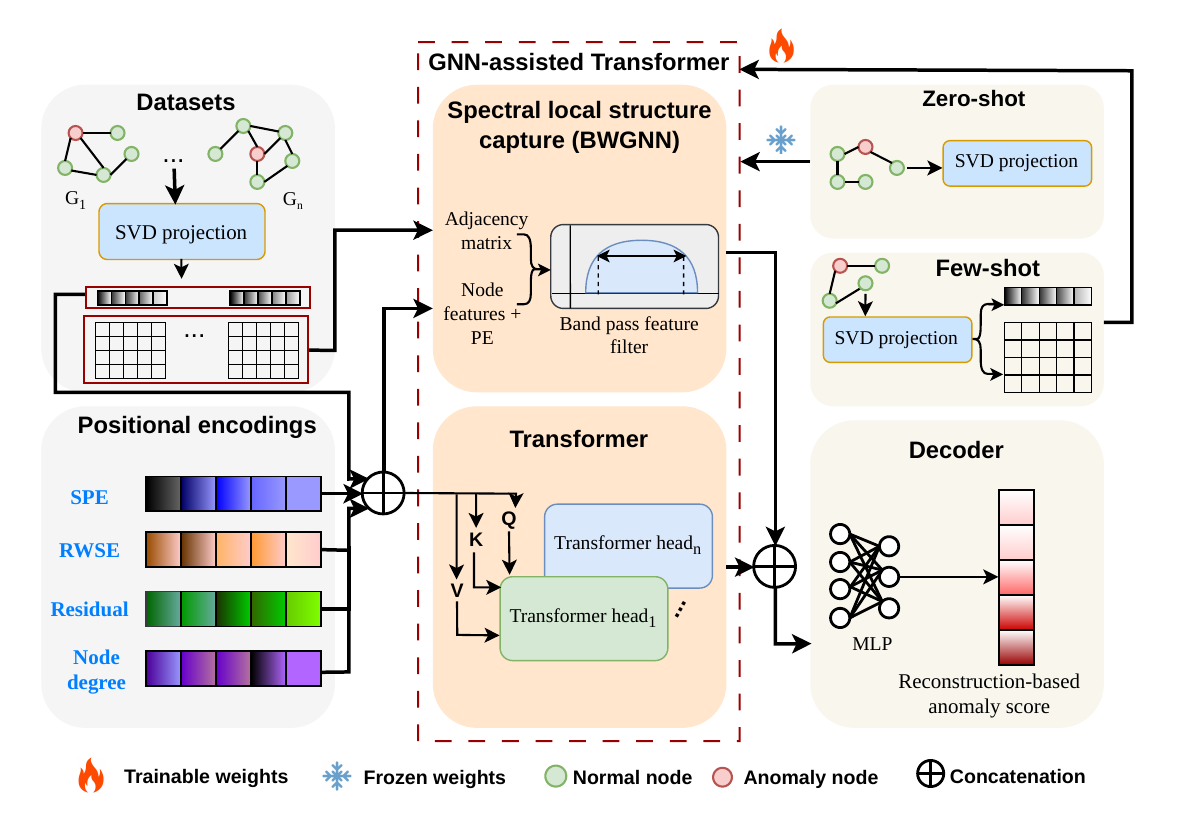}
    \caption{The framework of the proposed FoundAna.}
    \label{fig:schema}
\end{figure*}

\begin{definition}[Graph Anomaly Detection]
Given an attributed graph $\mathcal{G}$, graph anomaly detection aims to learn a scoring function $\phi: \mathcal{V} \rightarrow \mathbb{R}$ that assigns an anomaly score: $s_i = \phi(v_i) $ to each node $v_i \in \mathcal{V}$ (or $\mathcal{E}$, $\mathcal{G}$ for edge and graph anomaly, respectively). 
\end{definition}

\subsection{Graph Foundation Models}

\begin{definition}[Graph Foundation Model]
A graph foundation model is a pretrained encoder $\mathcal{F}_G$ that maps an attributed graph $\mathcal{G} = (\mathcal{V}, \mathcal{E}, \mathbf{X})$, with node feature matrix $\mathbf{X} \in \mathbb{R}^{|\mathcal{V}| \times d}$, to a domain-agnostic latent space: $\mathbf{Z} = \mathcal{F}_G(\mathcal{G};\, \theta^*)$, where $\theta^* = \arg\min_{\theta} \mathcal{L}_{\text{pre}}\left(\mathcal{F}_G(\cdot\,;\theta),\mathcal{G}\right)$ denotes the optimal parameters obtained by pretraining on a corpus of graphs, $\mathbf{Z} \in \mathbb{R}^{|\mathcal{V}| \times d_h}$ is the latent space node embedding matrix, and $d_h$ is the hidden dimension, such that $\mathbf{Z}$ generalizes across domains and tasks with minimal adaptation.
\end{definition}


A graph transformer treats each pre-processed node $u_i \in \mathcal{V}$ as a token and computes pairwise attention over all node pairs via the scaled dot-product mechanism. 
Formally, it can be defined as follows:

\begin{definition}[Graph Transformer]
A graph transformer is a function $\mathcal{T}: \mathbb{R}^{n \times d} \rightarrow \mathbb{R}^{n \times d_h}$ that operates on a set of node tokens $\{u_i\}_{i=1}^n$ without the canonical ordering assumption. 
Given an attributed graph $\mathcal{G} = (\mathcal{V}, \mathcal{E}, \mathbf{X})$, the input to the first layer is initialized as $\mathbf{H}^{(0)} = \mathbf{X} \in \mathbb{R}^{n \times d}$. 
At each layer $l$, given the node representation matrix $\mathbf{H}^{(l)} \in \mathbb{R}^{n \times d_h}$, the transformer computes projections 
$\mathbf{Q} = \mathbf{H}^{(l)}\mathbf{W}^Q$, 
$\mathbf{K} = \mathbf{H}^{(l)}\mathbf{W}^K$, 
$\mathbf{V} = \mathbf{H}^{(l)}\mathbf{W}^V$ with learnable weights 
$\mathbf{W}^Q, \mathbf{W}^K, \mathbf{W}^V \in \mathbb{R}^{d \times d_k}$, and produces an output via scaled dot-product attention:
\begin{equation}
\label{eqn:transformer}
\text{Attention}(\mathbf{Q}, \mathbf{K}, \mathbf{V}) = 
\text{softmax}\left(\frac{\mathbf{Q}\mathbf{K}^T}{\sqrt{d_k}}\right)\mathbf{V}
\end{equation}
where $d_k$ is the key dimension used for scaling. The attention output $\in \mathbb{R}^{n \times d_k}$ is subsequently projected to the hidden dimension via $\mathbf{W}^O \in \mathbb{R}^{d_k \times d_h}$, yielding the output node representations $\mathbf{H}^{(l+1)} \in \mathbb{R}^{n \times d_h}$. 
This attends to all $n \times n$ node pairs simultaneously and gives every node a global receptive field over the entire graph~\cite{ying2021transformers}.
\end{definition}

Unlike sequential text and vision-based transformers, self-attention treats the graph as a fully connected set of tokens, providing a global receptive field but discarding local structure inductive biases~\cite{ying2021transformers}. 
Therefore, the positional encoding in graph transformers augments $\mathbf{H}$ from graph topology prior to attention computation, restoring structural awareness without sacrificing global receptive fields, as discussed in Section~\ref{sec:GPE}.

\subsection{Problem Formulation}

\begin{definition}[Multi-Domain Graph Collection]

A multi-domain graph collection is defined as $\mathbb{G} = {\mathcal{G}^{(1)}, \mathcal{G}^{(2)}, \ldots, \mathcal{G}^{(M)}}$, where each $\mathcal{G}^{(m)} = (\mathcal{V}^{(m)}, \mathcal{E}^{(m)}, \mathbf{X}^{(m)})$ is an attributed graph from a distinct domain with potentially different feature dimensionality $d^{(m)}$, graph size $n^{(m)}$, and structural characteristics. 
\end{definition}

\begin{problem}[Foundation Model for Graph Anomaly Detection]
Given a multi-domain graph collection $\mathbb{G}$ with heterogeneous feature spaces $\mathbf{X}^{(m)} \in \mathbb{R}^{n^{(m)} \times d^{(m)}}$, our objective is to pretrain a unified encoder $\mathcal{F}_G$ with shared parameters $\theta^*$ and a decoder $\mathcal{D}$ that reconstructs both the node feature matrix $\hat{\mathbf{X}}^{(m)}$ and the adjacency matrix $\hat{\mathbf{A}}^{(m)}$ from the latent node embeddings $\mathbf{Z}^{(m)} = \mathcal{F}_G(\mathcal{G}^{(m)}; \theta^*)$. 
The anomaly score for each node $v_i$ is defined as the combined reconstruction error over its features and structural connections:
\begin{equation}
    s_i = \gamma_1 \cdot \left\| \mathbf{x}_i - \hat{\mathbf{x}}_i \right\|_2 + \gamma_2 \cdot \sum_{j \in \mathcal{N}(i)} \left| \mathbf{A}_{ij} - \hat{\mathbf{A}}_{ij} \right|
\end{equation}
where $\gamma_1, \gamma_2 \geq 0$ are weighting coefficients balancing feature and structural reconstruction errors. 
The model operates in an unsupervised setting and is designed to generalize across unseen graphs from new domains with minimal adaptation (e.g., zero-shot or few-shot settings).
\end{problem}

\section{Methodology: FoundAna}
In this section, we propose FoundAna, a universal GAD framework designed to operate across graph datasets without dataset-specific retraining (illustrated in Figure~\ref{fig:schema} and Algorithm~\ref{alg:foundana}).  
It consists of four stages: 
(1) SVD-based feature alignment (Section~\ref{sec:fea_align}), 
(2) multi-scale positional encoding (Section~\ref{sec:GPE}), 
(3) a GNN-assisted transformer that fuses local spectral and global attention representations (Section~\ref{sec:gnn_trans}), 
and (4) a dual reconstruction decoder for anomaly score (Section~\ref{sec:recon}).

\begin{algorithm}[t]
{\small
\caption{FoundAna: Training and Inference}
\label{alg:foundana}
\SetAlgoNoLine
\KwIn{A set of graphs $\{\mathcal{G}^{(m)}\}_{m=1}^{M}$, each with adjacency $\mathbf{A}^{(m)}$ and feature matrix $\mathbf{X}^{(m)} \in \mathbb{R}^{n^{(m)} \times d^{(m)}}$; unified dimension $d_u$; loss weights $\gamma_1, \gamma_2$; training epochs $T$}
\KwOut{Anomaly scores $\{s_i\}$ for all nodes at inference}

\For{each graph $\mathcal{G}^{(m)}$}{
    Compute truncated SVD: $\mathbf{X}^{(m)} \approx \mathbf{U}^{(m)} \boldsymbol{\Sigma}^{(m)} (\mathbf{V}^{(m)})^\top$\;
    Construct projection: $\mathbf{W}^{(m)} = \mathbf{V}^{(m)}_{1:d_u}$\;
    Obtain aligned features: $\tilde{\mathbf{X}}^{(m)} = \mathbf{X}^{(m)} \mathbf{W}^{(m)} \in \mathbb{R}^{n^{(m)} \times d_u}$\;
}

\For{each node $v_i$}{
    Compute $\text{SPE}(E_v, \lambda)$, $\text{RWSE}(v_i)$, residual $\mathbf{r}_i$, and $\log(1 + \deg(v_i))$\;
    Form descriptor: $\mathbf{p}_i = [\text{SPE} \| \text{RWSE} \| \mathbf{r}_i \| \log(1 + \deg(v_i))]$\;
    Form input token: $\mathbf{u}_i = [\tilde{\mathbf{x}}_i \| \mathbf{p}_i]$\;
}

\For{each training epoch $t = 1, \dots, T$}{
    Compute Beta-wavelet filters $\{F_B^{a, A-a}\}_{a=0}^{A}$ from $L$\;
    Aggregate: $\mathbf{h}_i^{\text{GNN}} = \sum_{a=0}^{A} F_B^{a,A-a} \mathbf{H}^{(a)}$\;
    Apply multi-head self-attention over tokens $\{\mathbf{u}_i\}$\;
    Obtain: $\mathbf{h}_i^{\text{Attn}}$ for all $v_i$\;
    $\mathbf{z}_i = [\mathbf{h}_i^{\text{Attn}} \| \mathbf{h}_i^{\text{GNN}}] \in \mathbb{R}^{2d_h}$\;

    Decode attributes: $\hat{\mathbf{x}}_i = \mathcal{D}_X(\mathbf{z}_i)$\;
    Decode adjacency: $\hat{\mathbf{A}}_{ij} =\mathcal{D}_A(\mathbf{z}_i,\mathbf{z}_j) = \sigma(\mathbf{z}_i^\top \mathbf{W}_A \mathbf{z}_j)$\;
    Compute $\mathcal{L}_X = \frac{1}{n}\sum_i \|\mathbf{x}_i - \hat{\mathbf{x}}_i\|_2^2$\;
    Compute $\mathcal{L}_A = -\frac{1}{n^2}\sum_{i,j}\left[\mathbf{A}_{ij}\log\hat{\mathbf{A}}_{ij} + (1-\mathbf{A}_{ij})\log(1-\hat{\mathbf{A}}_{ij})\right]$\;
    Update parameters by minimizing $\mathcal{L} = \gamma_1 \mathcal{L}_X + \gamma_2 \mathcal{L}_A$\;
}

\For{each node $v_i$}{
    $s_i = \gamma_1 \cdot \|\mathbf{x}_i - \hat{\mathbf{x}}_i\|_2 + \gamma_2 \cdot \sum_{j \in \mathcal{N}(i)} |\mathbf{A}_{ij} - \hat{\mathbf{A}}_{ij}|$\;
}
\uIf{graph-level inference}{
    $s_G = \sum_{i \in \mathcal{V}} \alpha_i \cdot s_i$\;
    \Return $s_G$\;
}
\Else{
    \Return $s_i$\;
}
}
\end{algorithm}

\subsection{SVD-based Feature Alignment}
\label{sec:fea_align}
To enable a single model to work on heterogeneous graph datasets with different node feature dimensions, we first project all node features into a shared latent space of dimension $d_u$. 
Formally, for each dataset with feature matrix $\mathbf{X^{(m)}} \in \mathbb{R}^{N^{(m)}\times d^{(m)}}$, we compute its truncated Singular Value Decomposition (SVD) $\mathbf{X}^{(m)} = \mathbf{U}^{(m)} \sum^{(m)}(\mathbf{V}^{(m)})^T$ and construct a dataset-specific projection matrix $\mathbf{W}^{(m)}= \mathbf{V}^{(m)}_{1:d_u} \in \mathbb{R}^{d^{(m)}\times d_u}$. 
The unified representation is then obtained as $\tilde{\mathbf{X}}^{(m)} = \mathbf{X}^{(m)} W^{(m)} \in \mathbb{R}^{N^{(m)} \times d_u}$, which gives low-dimension embedding that approximately preserves the dominant variance structure of the original features while aligning all datasets into a common feature space for downstream training and tasks. 

\subsection{Graph Positional Encoding}
\label{sec:GPE}

The goal of positional encoding (PE) is to augment each node $v_i$ with a structural descriptor $p_i \in \mathbb{R}^\text{p}$, such that the transformer input token for node $v_i$ becomes $u_i = [x_i || p_i]$, $u_i \in \mathbb{R}^{d_u + p}$, where $||$ denotes concatenation. 
To enhance the graph structural properties in anomaly detection, we selected four PE: SPE~\cite{huang2024stability}, RWSE~\cite{ma2023graph}, residual distance~\cite{qiao2025anomalygfm}, and node degree. 
SPE encodes stable and expressive global structural position and resolves the instability of prior methods (e.g., Laplacian PE) by replacing the hard partition of eigenspaces with a soft, eigenvalue-dependent partition~\cite{huang2024stability}. 
Given the $k$ smallest eigenpairs $(E_v,\lambda)$ of the normalized graph Laplacian defined by $L = I - D^{-1/2}AD^{-1/2}$, where $D$ is the diagonal degree matrix, SPE can be computed as: 
\begin{equation}
    SPE(V,\lambda)_i = \rho ( \sum_{l=1}^k \phi_l(\lambda_l). e_{vi}^{(l)},
\end{equation}
where $\phi_l: \mathbb{R} \rightarrow \mathbb{R}^r$ are learnable functions applied to the $l^{th}$ eigenvalue, $e_{vi}^{(l)}$ is the $l$-{th} eigenvector entry for node $v_i$, and $\rho: \mathbb{R}^r \rightarrow \mathbb{R}^p$ is a learnable aggregation. 

To characterize the local topological neighborhood of a node, we include RWSE as another PE that encodes the $k$-step random walk return probability for each node, represented by: 
\begin{equation}
    \text{RWSE}(v_i) = [(AD^{{-1}})^1_{ii}, (AD^{{-1}})^2_{ii}, \dots, (AD^{{-1}})^k_{ii}] \in \mathbb{R}^k
\end{equation}
where the $t$-{th} entry is the probability that a random walk starting from $v_i$ returns to $v_i$ after exactly $t$ steps. 
RWSE is entirely local and invariant to node labeling~\cite{rampavsek2022recipe}, therefore it can be defined for any graph without ambiguity.

In the GAD setting, the core inductive signal is that anomalous nodes exhibit atypical feature patterns relative to their neighbors, as shown in Figure~\ref{fig:res_dist}. 
To capture this feature-level deviation from the local neighborhood, we adopted the node representation residual introduced in~\cite{qiao2025anomalygfm} as the third encoding component. 
Instead of capturing the residuals in the GNN-derived node embeddings, we measure the residual in the raw node features, mathematically defined as:
\begin{equation}
    \text{r}_i = v_i - \frac{1}{|\mathcal{N}(v_i)|} \sum_{v_j \in \mathcal{N}(v_i)} v_j
\end{equation}
where $\mathcal{N}(v_i)$ denotes the neighbor set of $v_i$. 
Anomaly nodes generally have large residuals as they differ structurally or feature-wise from their context, compared with normal nodes. 

The degree of a node $\text{deg}(v_i) = \sum_j\mathbf{A}_{ij}$ provides a direct measure of local connectivity. 
As the node degree distribution is different for normal and anomalous nodes, as shown in Figure~\ref{fig:node_d}, we include the log-degree $\text{log}(1+\text{deg}(v_i))$ as a scalar appended to the feature vector. 
These four encoding components are concatenated to form the final PE for each node as follows: 
\begin{equation}
    \text{p}_i = [\text{SPE}(E_v,\lambda) || \text{RWSE}(v_i) || \text{r}_i || \text{log}(1+ \text{deg}(v_i))]
\end{equation}
where the dimension for $\text{p}_i$ is $\mathbb{R}^{\in \mathbb{R}^{\rho_{\text{SPE}}+ k + d_u + 1}}$ that is the sum of all PE length.

\subsection{GNN-assisted Transformer}
\label{sec:gnn_trans}
Inspired by GraphGPS~\cite{rampavsek2022recipe} architecture, we design a transformer backbone coupled with an anomaly-aware Beta wavelet filter-based GNN. 
The idea is to integrate both the local and global graph structural information by leveraging a GNN and a transformer, respectively. 

\noindent\textbf{GNN:} The GNN component captures the local structural information and high-frequency spectral patterns, a characteristic of an anomalous node~\cite{tang2022rethinking}. 
Tang et al.~\cite{tang2022rethinking} showed that the presence of anomalous nodes induces a rightward shift of graph-signal energy toward high Laplacian eigenvalues (described in Theorem~\ref{theorem:right_shift}), which suggests emphasizing spectrally localized band-pass filters over the traditional low-pass GNN (e.g., Graph Convolutional Network).

\begin{theorem}[Right-shift Aware Locality~\cite{tang2022rethinking}]
\label{theorem:right_shift}
Under the Gaussian anomaly model, the high-frequency area, $S_{high}(x)$, increases monotonically with anomaly degree, and Beta wavelet filters of sufficiently high order, $C$, can concentrate their spectral response on any interval $[\lambda_1,\lambda_2] \subset (0,2]$ while remaining $C$-hop localized in the graph.
\end{theorem}

To model this spectral information, we adopt the Beta-wavelet graph filters from BWGNN as the default local GNN encoder. 
Formally, the band pass filters $F$ can be obtained as:
\begin{equation}
F_B^{(\alpha,\beta)}(L) = U F(\Lambda) U^\top = \frac{\left(\frac{L}{2}\right)^\alpha \left(I - \frac{L}{2}\right)^\beta}{2 C(\alpha + 1, \beta + 1)},
\end{equation}
where $\alpha, \beta > 0$ controls the beta wavelet shape, $C(\alpha +1, \beta +1)$ is a normalization constant, $L$ denotes the normalized graph Laplacian with eigen-decomposition $L = U \Lambda U^T$, and $I$ is the identity matrix. 
During message passing, the SVD-projected node features $\mathbf{X}^{(0)}$ are filtered in parallel by all kernels in $F_B$, and the outputs are aggregated features as follows:
\begin{equation}
h_i^{GNN} = \sum_{a=0}^{A} F_B^{a,A-a} H^{(a)},
\end{equation}
where $A=\alpha + \beta$ controls the maximum scale, and $h_i^{GNN}$ is the graph encoding. 

\noindent\textbf{Transformer:}
Given the node tokens ${u_i}$, constructed by concatenating the SVD-projected features and PE, we apply a stack of standard transformer encoder layers with multi-head self-attention (shown in Eq.~\eqref{eqn:transformer}) and position-wise feed-forward networks. 
In contrast to the GNN, the transformer's global attention layers were trained without providing explicit edge features in the attention kernel but we relied on PE to include the topological information. 
Therefore, the transformer backbone produces node embeddings $h_i^{Attn}$, which capture the global, long-range dependencies across the graph~\cite{rampavsek2022recipe,min2022transformer}. 

Finally, we fuse the local and global information by concatenating the outputs of the GNN and transformer backbone at the node level, which produces the new features:
\begin{equation}
    \mathbf{z}_i = [\mathbf{h}^{\text{Attn}}_i || \mathbf{h}^{\text{GNN}}_i] \in \mathbb{R}^{2d_h},
\end{equation}
where $\mathbf{z}_i$ is the combined node embedding representation.

\subsection{Reconstruction and Anomaly Score Computation}
\label{sec:recon}
We employ separate decoders to reconstruct both node attributes and the graph adjacency structure from $\mathbf{z}_i$. 
For the attribute decoder $\mathcal{D_X}$, we use a two-layer Multi Layer Perceptron that maps $\mathbf{z}_i$ to the original feature space:
\begin{equation}
    \mathbf{\hat{x}}_i = \mathcal{D}_X(\mathbf{z}_i) = \sigma(\mathbf{W}_2 \cdot \text{ReLU}(\mathbf{W}_1\mathbf{z}_i + \mathbf{b}_1) + \mathbf{b}_2)
\end{equation}
where $\mathbf{W}_1 \in \mathbb{R}^{d_h \times 2d_h}$ compresses the fused representation to a bottleneck, and $\mathbf{W}_2 \in \mathbb{R}^{d^{(m)} \times d_h}$ maps back to the original feature dimensionality $d^{(m)}$ of graph $\mathcal{G}^{(m)}$. 
Similarly, since $\mathbf{z}_i$ consists of PE concatenated with node features, the latent representations of two connected nodes $v_i$ and $v_j$ implicitly encode their structural relationship. 
Edge existence can therefore be decoded via a bilinear decoder $\mathcal{D}_A$:
\begin{equation}
    \mathbf{\hat{A}}_{ij} = \mathcal{D}_A(\mathbf{z}_i,\mathbf{z}_j) = \sigma(\mathbf{z}_i^\top \mathbf{W}_A \mathbf{z}_j)
\end{equation}
where $\mathbf{W}_A \in \mathbb{R}^{2d_h \times 2d_h}$ is a learnable bilinear weight matrix. 
The overall reconstruction loss is defined as:
\begin{equation}
    \mathcal{L} = \gamma_1 \mathcal{L}_X + \gamma_2 \mathcal{L}_A,
\end{equation}
where $\gamma_1, \gamma_2 \geq 0$ balance the two reconstruction objectives, and the feature and structure reconstruction losses are respectively given by:
\begin{align}
    \mathcal{L}_X &= \frac{1}{n} \sum_{i=1}^n \|\mathbf{x}_i - \mathbf{\hat{x}}_i\|_2^2, and \\
    \mathcal{L}_A &= -\frac{1}{n^2} \sum_{i=1}^n \sum_{j=1}^n \left[ \mathbf{A}_{ij} \log \mathbf{\hat{A}}_{ij} + (1 - \mathbf{A}_{ij}) \log(1 - \mathbf{\hat{A}}_{ij}) \right],
\end{align}

In inference, a node is considered anomalous if it cannot be reconstructed in either attribute or structural space.
We therefore define a composite anomaly score for node $v_i$ in Eq.~\eqref{eq:node-a}.
For graph-level anomaly detection, a readout layer aggregates node attributes and adjacency reconstruction errors through attention-weighted pooling, which produces a single graph-level anomaly score that reflects the collective structural and feature deviation across the graph in Eq.~\eqref{eqn:readout}.

\begin{equation}
\label{eq:node-a}
   s_i = \gamma_1 \cdot \|\mathbf{x}_i - \mathbf{\hat{x}}_i\|_2 + \gamma_2 \cdot \sum_{j \in \mathcal{N}(i)} \left| \mathbf{A}_{ij} - \mathbf{\hat{A}}_{ij} \right|
\end{equation}
 
\begin{equation}
\label{eqn:readout}
     s_G = \text{Readout}\left(\{s_i\}_{i \in \mathcal{V}}\right) = \sum_{i \in \mathcal{V}} \alpha_i \cdot s_i
\end{equation}

\section{Experiments}
We validate the proposed FoundAna framework through experiments addressing the following Research Questions (RQs):
\begin{itemize}
    \item \textbf{RQ1}: Does the proposed framework outperform baselines across evaluated datasets?
    \item \textbf{RQ2}: Does the model generalize to unseen graph data under zero-shot and few-shot scenarios?
    \item \textbf{RQ3}: Does concatenating PE with raw features improve GAD performance over using raw features alone?

\end{itemize}

\subsection{Datasets}
We evaluate on 9 GAD benchmark datasets summarized in Table~\ref{tab:dataset_info}, which span social networks, financial and e-commerce networks, and academic citation graphs.

\noindent\textbf{Social networks:} The Question dataset is derived from Community Question Answering (CQA) platforms and represents a bipartite user–question graph, where anomalies correspond to malicious users or low-quality automated content~\cite{qiao2025deep}. 
Weibo captures user–post interactions from a Chinese microblog platform, commonly used to detect Sybil accounts and bot-driven rumor propagation. 
Facebook maps ego-network social connections and profile features, originally introduced for social circle discovery~\cite{qiao2025deep}. 
Reddit captures cross-subreddit user interactions, with anomalies defined by outlier engagement patterns deviating from the regular user base~\cite{qiao2025deep}. 

\noindent\textbf{Financial and e-commerce networks:} YelpChi is a multi-relational review graph from the Chicago area, where fraudulent reviewers are labeled as anomalies via filtering heuristics~\cite{qiao2025deep}. 
Amazon is an e-commerce platform targeting fraudulent reviewers who artificially inflate product ratings~\cite{qiao2025deep}. 
T-Finance is a transaction network designed for detecting fraudulent loan accounts with atypical topological patterns~\cite{tang2022rethinking}. 
DGraph is a massive real-world fintech dataset with millions of user nodes and financial or social ties, specifically targeting fraudulent loan applications~\cite{qiao2025deep}. 
Elliptic is a large-scale Bitcoin transaction graph where illicit transactions are identified by association with known criminal entities~\cite{qiao2025deep}.

\noindent\textbf{Citation network:} Ogbn-arxiv is a directed Computer Science (CS) paper citation network, where anomalies represent structural outliers or misclassified papers within the citation hierarchy~\cite{qiao2025deep}.

\begin{table*}[t]
    \centering
        \caption{Description of the dataset. T-finance is primarily used for training, whereas the rest of the dataset is used for testing.}
    \begin{tabular}{llcccccc}
    \toprule
    \textbf{Dataset} & 
    \textbf{Domain} & 
    \textbf{Anomaly percentage} & 
    \textbf{Dimension}&
    \textbf{Nodes} & 
    \textbf{Edges}  & 
    \textbf{Normal degree} & 
    \textbf{Anomaly degree} \\  
    
    \midrule
    T-Finance & Finance  & 4.580 & 10  & 39,357 & 42,445,086 & 1098.960 & 651.460 \\
    \midrule
    Question & Q\&A platform & 2.980 & 301 & 48,921 & 202,461  & 3.960 & 9.840  \\
    
    YelpChi & Reviews & 5.110 & 32 & 23,831 & 98,630 & 4.280 & 1.590 \\
    
    Weibo & Social media & 10.330 & 400 & 8,405 & 754,542 & 95.240 & 42.340 \\
    
    Dgraphfin & Finance  & 67.300 & 17 & 3,700,550 & 4,300,999 & 1.730 & 0.890 \\

    Facebook & Social network & 2.310 & 128 & 1,081 & 55,104 & 52.060 & 5.280 \\

    
    Amazon & Reviews & 6.780 & 25 & 10,224 & 351216 & 32.590 & 58.590 \\
    
    Ogbn-arixv & Citation network & 3.540 & 128 & 169,343 & 2,357,596  & 13.680 & 20.500 \\
    
    Elliptic & Finance  & 9.760 & 165 & 46,564 & 73,248  & 1.660 & 0.810 \\
    
    Reddit & Social media  & 3.330 & 64 & 10,984 & 168,016  & 15.400 & 12.380 \\


    \bottomrule
    \end{tabular}

    \label{tab:dataset_info}
\end{table*}

\subsubsection{Node and Edge Duality Assumption}
Since none of these datasets consists of the edge features, we consider a common anomaly prediction for node and edge-based tasks. 
To support this assumption, we use a graph-theoretic perspective, where node and edge anomalies are unified under a common representational framework through the concept of the line graph transformation (Definition ~\ref{def:lft}).

\begin{definition} [Line Graph Transformation~\cite{whitney1932congruent, bondy2008graph}]
\label{def:lft}
Given a graph $\mathcal{G}$, its line graph $L(\mathcal{G})$ is constructed such that each edge $e \in \mathcal{E}$ in the original graph becomes a node $v' \in \mathbf{V'}$ in $L(\mathcal{G})$, with two nodes in $L(\mathcal{G})$ connected if their corresponding edges in $\mathcal{G}$ share a common endpoint. 
This makes detecting anomalous edges in $\mathcal{G}$ mathematically equivalent to detecting anomalous nodes in $L(\mathcal{G)}$, which establishes a formal duality between node-level and edge-level GAD.
\end{definition}


\subsection{Baselines and Metrics}
We evaluate the proposed FoundAna against two categories of state-of-the-art methods: GNN-based GAD and GFM-based GAD. 
The GNN-based baselines include GCN \cite{kipf2016semi}, BWGNN \cite{tang2022rethinking}, DOMINANT \cite{ding2019deep}, AnomalyDAE \cite{fan2020anomalydae}, and CoLA \cite{liu2021anomaly}. 
These methods have various design choices, ranging from low-pass spectral filters and graph autoencoders to contrastive self-supervised learning, representing the landscape of GAD. 
The GFM baselines include ARC \cite{liu2024arc}, AnomalyGFM \cite{qiao2025anomalygfm}, and UNPrompt \cite{niu2024zero}, each of which aims at cross-graph generalization under different pretraining and prompting approaches.

Since GAD datasets are inherently class-imbalanced, we select three complementary evaluation metrics: AUROC, which measures ranking quality independent of threshold; AUPRC, which is sensitive to performance on the minority anomaly class; and Macro F1, which evaluates per-class balance after thresholding. 
Also, to assess the statistical significance of the performance, we employ the Wilcoxon signed-rank test~\cite{wilcoxon1945individual}, a non-parametric statistical test that evaluates whether the differences between paired observations are systematically different from zero without assuming a normal distribution of the data. 
We consider the performance metrics on individual data as statistically significant results when $p<$ 0.05.

\subsection{Implementation Details}
All experiments are implemented in Python v3.10.20, leveraging PyTorch v2.4.0, PyTorch Geometric v2.6.0, PyTorch Scatter v2.1.2, and scikit-learn v1.7.2. 
Training and inference for both the proposed model and all baselines were conducted on a single NVIDIA L40S GPU with 48 GB of memory. 
Since most of the baselines used a single dataset to train the models, and T-Finance being the widely adopted large-scale benchmark for anomaly detection in the GAD literature~\cite{qiao2025anomalygfm, tang2022rethinking}, we selected T-Finance to train the model and evaluated zero-shot and few-shot on 9 anomaly detection datasets.
Also, it comprises a huge network of connections with real-world fraud annotations.
The weights are frozen when inferring in a zero-shot setting, whereas they are fine-tuned when inferring in a few-shot setting.

The final configuration uses 8 attention heads, a batch size of 2,048, and 100 training epochs with a learning rate of $1 \times 10^{-3}$. 
Feature unification via SVD projects all node attributes to a common dimension of 32. 
For PEs, SPE uses the 8 non-trivial Laplacian eigenpairs, and RWSE is computed with 16 random walk steps. 
The reconstruction loss is a weighted combination of the feature reconstruction and adjacency reconstruction terms, with loss weights $\gamma_1 = \gamma_2 = 0.5$, which gives equal importance assigned to attribute fidelity and structural reconstruction. 
For the few-shot inference, we use 20 data samples to train the model for 30 epochs.


\begin{table*}[tbh]
\centering
\renewcommand{\arraystretch}{1.1}
\setlength{\tabcolsep}{4pt}
\caption{Performance comparison across datasets using AUROC, AUPRC, and Macro F1 scores. The numbers indicate results when training only on the T-finance dataset.   
Avg. is the average performance on all nine benchmark datasets.
\textbf{Bold} indicates the best result per column; \underline{underline} indicates the second best.}
\begin{tabular}{clccccccccc |cc}
\toprule
\multirow{2}{*}{\textbf{Metrics}} & \multirow{2}{*}{\textbf{Methods}} 
  & \multicolumn{9}{c}{\textbf{Dataset}} 
  & \multirow{2}{*}{\textbf{Avg.}} & \multirow{2}{*}{\textbf{p-value}}  \\
\cmidrule(lr){3-11}
& & Question & YelpChi & Weibo & Dgraphfin & Facebook & Amazon & Ogbn-arxiv & Elliptic & Reddit 
&  & \\
\midrule

\multirow{8}{*}{\rotatebox{90}{AUROC}} & GCN & 0.434  & 0.592   & 0.227  & 0.434  &0.406  &0.447  & 0.421  & 0.442  & 0.453  
& 0.431 & 0.003\\
& BWGNN & 0.492 & 0.642 & 0.204 & 0.505 & 0.401 & 0.602 & 0.456 & 0.462 & 0.506 
& 0.474 & 0.003 \\
& Dominant & 0.533 & 0.717 & 0.767 & OOM &0.576 & 0.585 & OOM & 0.307 & 0.503 
& 0.569 & 0.031 \\
& AnomalyDAE & 0.519 & 0.404 & 0.499 & OOM & 0.023 & 0.397 & 0.425 & 0.448 & 0.421 
& 0.392 & 0.007 \\
& CoLA & 0.489 & 0.400 & 0.388 & \underline{0.555} & 0.515 & 0.513 & 0.564 & 0.501 & 0.490 
& 0.490 & 0.003 \\

\cmidrule{2-13}
& ARC    & \underline{0.572} & 0.488 & \textbf{0.877} & 0.498 & 0.643 & 0.609 & \textbf{0.814} & 0.257 & \underline{0.571}
& 0.584  & 0.570\\
& AnomalyGFM    & 0.546  & 0.524 & 0.556 & OOM&\textbf{0.733} & 0.511 & OOM & \textbf{0.606} & 0.533 
& \underline{0.572} & 0.375 \\
& UNPrompt    &0.468 &0.522 & \underline{0.854} &0.419 &\underline{0.732} &\underline{0.661} & \underline{0.759} & 0.262 & 0.534
& 0.568 & 0.359 \\
& FoundAna-zero shot & 0.477 & 0.566 & 0.749 & 0.364 & 0.483 & 0.521 & 0.589 & 0.466 & 0.512 &0.525 & - \\
& FoundAna-few shot & \textbf{0.585} & \textbf{0.915} & 0.801 & \textbf{0.560} & 0.546 & \textbf{0.695} & 0.689 & \underline{0.508} & \textbf{0.586} & \textbf{0.647} & - \\
\midrule

\multirow{8}{*}{\rotatebox{90}{AUPRC}} & GCN & 0.024 & 0.087 &0.236  & 0.598  &0.019  & 0.068  & 0.028 & 0.023 &0.035 
& 0.124 & 0.003 \\
& BWGNN & 0.027 & 0.098 & 0.064 & 0.780 & 0.097  & 0.101 & 0.030 & 0.120 & 0.038 
& 0.150 & 0.300 \\
& Dominant & 0.032 & \underline{0.166} & 0.481 & OOM & 0.108 & 0.083 & OOM & 0.064 & 0.035 
& 0.138 & 0.109 \\
& AnomalyDAE & 0.031 & 0.040 & 0.130 & OOM & 0.012 & 0.063 & 0.029 & 0.083 & 0.028 
& 0.052 & 0.007\\
& CoLA & 0.033 & 0.048  & 0.231  & \textbf{0.715} & 0.033  & 0.077  & 0.052 & 0.091 & 0.034 
& 0.146 & 0.054 \\

\cmidrule{2-13}
& ARC    & \underline{0.039} & 0.052 & \textbf{0.618} & 0.673 & 0.098 & 0.090 & \textbf{0.285} & 0.067 & \underline{0.040} 
& \underline{0.218} & 0.652\\
& AnomalyGFM    & 0.034& 0.060 & 0.113 & OOM &\textbf{0.115} & 0.063 & OOM & \underline{0.127}& 0.034 
& 0.078 & 0.156 \\
& UNPrompt    & 0.031& 0.058 & 0.491 & 0.608 & \underline{0.113} & \underline{0.106} & \underline{0.247} & 0.062 & 0.036 
& 0.177 & 0.250 \\
& FoundAna-zero shot & 0.028 & 0.146 & 0.557  & 0.586  & 0.042 & 0.068 & 0.042 & 0.083 & 0.037 & 0.176 & -\\
& FoundAna-few shot & \textbf{0.041} & \textbf{0.324} & \underline{0.601} & \underline{0.682} & 0.048 & \textbf{0.122} & 0.050 & \textbf{0.132} & \textbf{0.044} & \textbf{0.227}  & - \\
\midrule

\multirow{8}{*}{\rotatebox{90}{Macro F1}} & GCN & 0.455  & 0.490 & 0.548  & 0.441  & 0.445 & 0.480 & 0.436 &0.525 & 0.479  
& 0.476 & 0.003 \\
& BWGNN & 0.488  & 0.529 & 0.474 & 0.427 & 0.495 &0.504 & 0.485 & 0.478 &0.502 
& 0.486 & 0.003 \\
& Dominant & 0.490 & 0.486 & 0.551 & OOM & 0.493 & 0.504 & OOM & 0.476 & 0.459 
& 0.494 & 0.015 \\
& AnomalyDAE & 0.455 & 0.480 & 0.088 & OOM & 0.494 & 0.467 & 0.148 & 0.474 & 0.475 
& 0.385 & 0.007
\\
& CoLA & 0.472 & 0.451 & 0.517 & 0.450 & 0.484 & 0.484 & 0.472 & 0.479 & 0.481 
& 0.471 & 0.003\\

\cmidrule{2-13}
& ARC    & 0.497 & 0.486 & 0.095 & \underline{0.496} & 0.356 & 0.482 & 0.491 & 0.258 & 0.491 
& 0.368 & 0.007 \\
& AnomalyGFM    & 0.403& 0.331 & 0.413 & OOM & \underline{0.519} & 0.347 & OOM & \textbf{0.580}& 0.260 
& 0.407 & 0.078 \\
& UNPrompt    & \underline{0.520} & 0.514 & 0.722 & 0.494 & \textbf{0.592} & \underline{0.537} & \textbf{0.653} & 0.474 & \underline{0.503} 
& \underline{0.550} & 0.425 \\
& FoundAna-zero shot & 0.497  & \underline{0.540} & \underline{0.728} & 0.427 & 0.488  & 0.486  & 0.490 & 0.454 & 0.496  & 0.511 & - \\
& FoundAna-few shot & \textbf{0.533} & \textbf{0.693} & \textbf{0.740} & \textbf{0.499} & 0.508 & \textbf{0.583}& \underline{0.502} & \underline{0.528} & \textbf{0.532} & \textbf{0.568}& -\\
\bottomrule

\end{tabular}

\label{tab:results}
\end{table*}

\subsection{Results of FoundAna}
Table~\ref{tab:results} illustrates the performance comparison of the baseline and FoundAna on the nine benchmark datasets under three evaluation metrics. 
The rightmost columns report the average performance across all datasets and the $p$-value from the Wilcoxon signed-rank test to assess statistical significance. 
The bold values denote the best performance per dataset, and underlined values indicate the second best. 
Out of Memory (OOM) entries refer to methods that failed due to memory constraints, which generally occur on larger datasets such as Dgraphfin and Ogbn-arxiv of some baselines, indicating their intensive computation cost. 

Generally, our proposed method achieves the highest average performance across all three metrics under the few-shot inference with AUROC, AUPRC, and Macro F1 scores of 0.647, 0.227, and 0.568, respectively. 
Also, our method has a clear superiority over the GNN-based baselines. 
For instance, our method has higher performance than the best AUROC and Macro F1 of 0.569 and 0.494, respectively, achieved in Dominant and AUPRC of 0.150 in BWGNN. 
Similarly, we observe comparable or superior results on the GFM baselines. 
A higher performance on all three metrics is observed on the four datasets (i.e., Question, YelpChi, Amazon, and Reddit) under a few-shot learning setting. 
Moreover, the highest gains are observed on YelpChi (AUROC 0.915, Macro F1 0.693), Weibo (Macro F1 0.740), and Dgraphfin (AUPRC 0.682), which support an affirmative answer to RQ1. 

Also, \textit{FoundAna-zero shot already achieves competitive generalization with an average AUROC of 0.525, AUPRC of 0.176, and Macro F1 of 0.511, frequently surpassing fully supervised GNN baselines without any labeled data at test time}. 
This supports that FoundAna generalizes to unseen graph data under zero-shot and few-shot scenarios (RQ2). 
The performance superiority is statistically significant for all GNN-based baselines, except BWGNN for AUPRC (i.e., $p$-value of 0.300); however, for the GFM baseline, only the Macro F1 score on ARC is statistically significant.

\begin{figure*}[tbh]
    \centering
    \includegraphics[width=\linewidth]{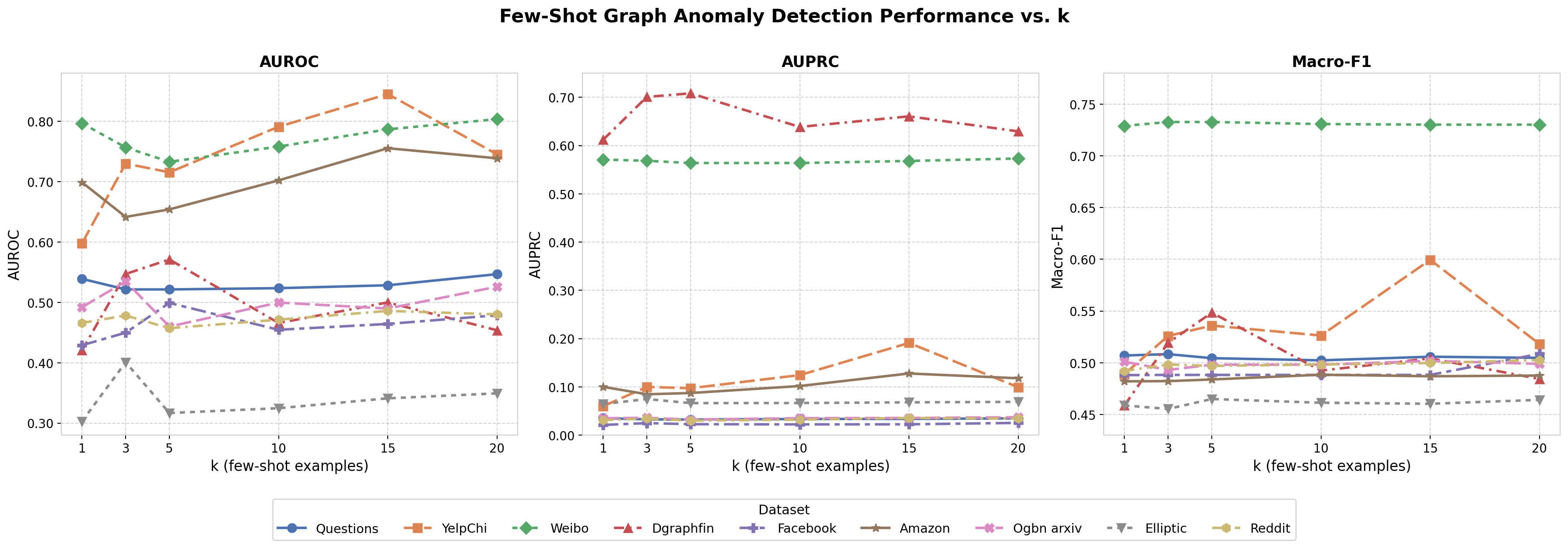}
    \caption{Performance results on nine datasets under different values of $k$ in few-shot learning settings.}
    \label{fig:few_shot}
\end{figure*}

Figure~\ref{fig:few_shot} illustrates the few-shot performance of FoundAna as a function of the number of few-shot examples $k \in \{1, 3, 5, 10, 15, 20\}$ across nine datasets under AUROC, AUPRC, and Macro F1. 
Weibo consistently achieves the strongest and most stable performance across all $k$ values (e.g., AUROC $\approx$ 0.790–0.810), while YelpChi exhibits a clear upward trend, peaking at AUROC $\approx$ 0.840 and Macro F1 $\approx$ 0.60 at $k$=15. 
Dgraphfin achieves notably high AUPRC values peaking near 0.70 at $k$=5, whereas the remaining datasets cluster in lower performance ranges.

\subsection{Ablation Study}
In this section, we discuss the contributions of various components of the proposed framework, specifically the effect of different PEs used, the performance comparison across three GNN backbones, and the performance when using different types of features in the foundation model.  

\subsubsection{Effect of PEs}
Table~\ref{tab:abla_pe} ablates the contribution of each PE component on the YelpChi dataset (results on other datasets are in the supplementary material). 
The full combination of all PEs achieves the best performance across AUROC (0.804), AUPRC (0.165), and Macro F1 (0.581). 
Removing RWSE causes the smallest drop, whereas removing RD and ND leads to the sharpest decline of AUROC to 0.612 and 0.660, respectively. 
This suggests that RWSE carries partially redundant information with other PEs, while RD and NE are important for distinguishing anomalous nodes. 
Also, we observed poor performance when none of the PEs were used.

\begin{table}[t]
    \centering
     \caption{Results when using different combinations of PE. 
     The reported results are on the YelpChi dataset (other datasets are in the supplementary material). PE: Positional Encoding; SPE: Stable and Expressive Positional Encoding; RWSE: Random Walk Structural Encoding;
     RD: Residual Distance; ND: Node Degree.}
    \begin{tabular}{l cccc}
     \toprule
     PE & AUROC & AUPRC & Macro F1 & Training time \\
     \midrule
     No PE & 0.634 & 0.088 & 0.537 & \textbf{0.160 }\\
     w/o SPE & 0.745 & 0.118 & 0.546 & 0.230 \\
     w/o RWSE & 0.764 & 0.149 & 0.580 & 0.220 \\
     w/o RD & 0.612 & 0.085 & 0.536 & 0.210 \\
     w/o ND & 0.660 & 0.082 & 0.526 & 0.230 \\
     w/ all PE & \textbf{0.804} & \textbf{0.165} &\textbf{0.581} & 0.220 \\
\bottomrule
    \end{tabular}
   
    \label{tab:abla_pe}
\end{table}

\subsubsection{Effect of GNN Backbone Selection}
Table ~\ref{tab:abla_gnn} compares three GNN backbone choices on the Ogbn-arxiv dataset (results on other datasets are in the supplementary material). 
BWGNN achieves the best overall performance on AUROC (0.546) and Macro F1 (0.496), which is consistent with prior work showing that BWGNN's spectral filtering is particularly effective at capturing anomalous patterns~\cite{tang2022rethinking,zhang2025heterogeneous}. 
GCN and Graph Isomorphism Network (GIN) perform comparably but fall short of BWGNN, whereas the best AUPRC (0.039) is observed when completely removing the GNN backbone. 
This indicates that general-purpose message-passing backbones are suboptimal for anomaly-specific tasks in our framework.

\begin{table}[t]
    \centering
    \scriptsize
     \caption{Performance results on different GNN backbones when tested on the ogbn-arixv dataset.
     T: Training; I: Inference.}
    \begin{tabular}{l cccc}
     \toprule
     Model & AUROC & AUPRC & Macro F1 & T/I time \\
     \midrule
     GCN & 0.419 & 0.028 & 0.491 & 318.100/ 3.350 \\
     GIN & 0.301 & 0.024 & 0.489 & \textbf{304.800}/ 3.140 \\
     BWGNN & \textbf{0.546} & 0.037 & \textbf{0.496} & 465.100/ 4.870 \\
     w/o GNN & 0.334 & \textbf{0.039} & 0.386 & 305.300/ \textbf{2.300} \\
\bottomrule
    \end{tabular}
   
    \label{tab:abla_gnn}
\end{table}

\subsubsection{Effect of PE-Augmented Input Features on GAD}
Table~\ref{tab:abla_gnn_pe} evaluates whether augmenting raw features with PE benefits BWGNN across three representative datasets from each dataset group: social networks (Weibo), financial and e-commerce (YelpChi), and citation (Ogbn-arxiv) networks. 
For this experiment, we select only BWGNN as it performs the best compared to all other GNN models, as presented in Table~\ref{tab:abla_gnn}. 

We observe a dataset-dependent pattern: on YelpChi and Ogbn-arxiv, the raw feature combined with PE consistently outperforms the raw feature alone, with an AUROC gain of 0.103 and 0.085 on YelpChi and Ogbn-arxiv, respectively. 
However, on Weibo, raw feature slightly outperforms the case of PE combination, with an AUROC of 0.006, and marginally achieves better on AUPRC and Macro F1. 
This indicates that in the dense social networks, raw feathers already capture sufficient discriminative signals and PE only contributes marginally, which is consistent in other datasets as well. 
These results provide an answer to our \textbf{RQ3}: PE-augmented features generally improve AD, though the magnitude of improvement is domain-dependent.

\begin{table}[t]
    \centering
    \scriptsize
    \caption{Performance results when input is raw features and PE concatenated features (Full results are in supplementary material).
    PE: Positional Encoding; RF: Raw Features; T: Training; I: Inference.}
    \begin{tabular}{l cccc}
     \toprule
     Model & AUROC & AUPRC & Macro F1 & T/I time \\
     \midrule
     \multicolumn{5}{c}{\textit{Weibo}} \\
     \midrule
     BWGNN (RF) &  \textbf{0.758} & 0.562   &  0.730  &  \textbf{760.9}/ \textbf{0.410}     \\
     BWGNN (RF + PE)  & 0.752 & \textbf{0.564}  & \textbf{0.732}   & 769.1/ 0.440      \\
    \midrule
     \multicolumn{5}{c}{\textit{YelpChi}} \\
     \midrule
     BWGNN (RF) & 0.563 & 0.067  & 0.521   &  \textbf{760.9}/ \textbf{1.220}     \\
     BWGNN (RF + PE)  & \textbf{0.666} & \textbf{0.082}  &  \textbf{0.526} &  769.1/ 1.230    \\
    \midrule
     \multicolumn{5}{c}{\textit{Ogbn Arxiv}} \\
     \midrule
     BWGNN (RF) & 0.496 & 0.035  & \textbf{0.497}   & \textbf{760.9}/ 10.690       \\
     BWGNN (RF + PE)  & \textbf{0.581} & \textbf{0.039} & 0.495 & 769.1/ \textbf{10.630}  \\
\bottomrule
    \end{tabular}
    \label{tab:abla_gnn_pe}
\end{table}

\section{Conclusion}
\label{sec:con}

In this work, we investigated the graph anomaly detection problem through the lens of foundation models. 
Specifically, we introduced FoundAna, which integrates positional encoding with anomaly-specific GNNs to produce structural representations that generalize across diverse graph domains. 
The experimental results demonstrate that FoundAna consistently achieves better average results across 9 GAD benchmarks, validating its effectiveness as a graph-agnostic framework. 
Moreover, we have conducted extensive ablation studies on PE combinations and GNN backbone selection, revealing the critical role of PE in capturing anomalous patterns across heterogeneous graph domains. 
Nevertheless, the applicability of FoundAna to knowledge graphs remains an open challenge, representing a promising direction for future work in extending foundational GAD frameworks to more complex, heterogeneous graph structures.

\bibliographystyle{IEEEtran}   
\bibliography{references}       


\end{document}